%% file: main.tex
\documentclass{article}

\usepackage[preprint]{corl_2024} 
\usepackage{booktabs}
\usepackage{array}
\usepackage{rotating}
\usepackage{graphicx}
\usepackage{tabularx}
\usepackage{adjustbox}
\usepackage{makecell}
\usepackage{tabularray}
\usepackage{diagbox}
\usepackage{colortbl}
\usepackage{multirow}
\usepackage{amsmath}
\usepackage{amssymb}
\usepackage{algorithm}
\usepackage{algorithmic}
\usepackage{wrapfig}
\usepackage{caption}
\usepackage{enumitem}
\input{definitions}
\title{Lift, Splat, Map: Lifting Foundation Masks for Label-Free Semantic Scene Completion}

\author{
  Arthur Zhang\\
  Department of Computer Science\\
  University of Texas at Austin\\
  \texttt{arthurz@cs.utexas.edu} \\
  \And
  Rainier Heijne\\
  Department of Mechanical Engineering\\
  Eindhoven University of Technology\\
  \texttt{r.f.p.heijne@student.tue.nl} \\
  \And
  Joydeep Biswas\\
  Department of Computer Science\\
  University of Texas at Austin\\
  \texttt{joydeepb@cs.utexas.edu} \\
}

\begin{document}
\maketitle


\begin{abstract}
Autonomous mobile robots deployed in urban environments must be \textit{context aware}, i.e, able to distinguish between different semantic entities, and robust to occlusions. Current approaches like semantic scene completion  (SSC) require pre-enumerating the set of classes and costly human annotations, while representation learning methods relax these assumptions but are not robust to occlusions and learn representations tailored towards auxiliary tasks. To address these limitations, we propose \ourmodel{}, a method that lifts masks from visual foundation models to predict a continuous, open-set semantic and elevation-aware representation in bird's eye view (BEV) for the entire scene, including regions underneath dynamic entities and in occluded areas. Our model only requires a single RGBD image, does not require human labels, and operates in real-time. We quantitatively demonstrate our approach outperforms existing models trained from scratch on semantic and elevation scene completion tasks with finetuning. Furthermore, we show that our pre-trained representation outperforms existing visual foundation models at unsupervised semantic scene completion. We evaluate our approach using \coda{}, a large-scale, real-world urban robot dataset. Supplementary visualizations, code, data, and pre-trained models, will be publicly available soon.
\end{abstract}




\input{introduction}

\input{relatedwork}

\input{approach}

\input{implementation}

\input{experiments}

\input{limitations}

\input{conclusion}

	



\clearpage




\bibliography{refs}  

\clearpage

\input{appendix}

\end{document}

%% file: definitions.tex
\definecolor{Orange}{rgb}{0.9608, 0.5882, 0.3922}
\definecolor{Yellow}{rgb}{0.9608, 0.9020, 0.3922}
\definecolor{Brown}{rgb}{0.5882, 0.2353, 0.1176}
\definecolor{Maroon}{rgb}{0.7059, 0.1176, 0.3137}
\definecolor{Grapefruit}{rgb}{1, 0.3137, 0.3922}
\definecolor{Royal Blue}{rgb}{0.1176, 0.1176, 1}
\definecolor{Magenta}{rgb}{0.7843, 0.1569, 1}
\definecolor{Pink}{rgb}{1, 0.5882, 1}
\definecolor{Purple}{rgb}{0.2941, 0, 0.2941}
\definecolor{White}{rgb}{1, 1, 1}
\definecolor{Black}{rgb}{0, 0, 0}
\definecolor{Light Green}{rgb}{0.3411, 0.9764, 0.4274}
\definecolor{Light Red}{rgb}{0.9764, 0.3411, 0.3411}
\definecolor{Yellow}{rgb}{1, 0.9294, 0.4784}

\definecolor{unknownColor}{RGB}{0, 0, 0}
\definecolor{concreteColor}{RGB}{97, 171, 47}
\definecolor{grassColor}{RGB}{159, 77, 200}
\definecolor{rocksColor}{RGB}{141, 49, 126}
\definecolor{speedwayBricksColor}{RGB}{235, 128, 55}
\definecolor{redBricksColor}{RGB}{174, 149, 8}
\definecolor{pebblePavementColor}{RGB}{98, 3, 141}
\definecolor{lightMarbleTilingColor}{RGB}{74, 110, 203}
\definecolor{darkMarbleTilingColor}{RGB}{115, 240, 49}
\definecolor{dirtPathsColor}{RGB}{127, 57, 78}
\definecolor{roadPavementColor}{RGB}{142, 143, 60}
\definecolor{shortVegetationColor}{RGB}{17, 187, 187}
\definecolor{porcelainTileColor}{RGB}{165, 247, 137}
\definecolor{metalGratesColor}{RGB}{27, 183, 89}
\definecolor{blondMarbleTilingColor}{RGB}{80, 29, 134}
\definecolor{woodPanelColor}{RGB}{244, 81, 150}
\definecolor{patternedTileColor}{RGB}{159, 77, 163}
\definecolor{carpetColor}{RGB}{116, 100, 60}
\definecolor{crosswalkColor}{RGB}{153, 207, 156}
\definecolor{domeMatColor}{RGB}{159, 138, 135}
\definecolor{stairsColor}{RGB}{131, 217, 44}
\definecolor{doorMatColor}{RGB}{131, 97, 123}
\definecolor{thresholdColor}{RGB}{101, 226, 115}
\definecolor{metalFloorColor}{RGB}{40, 43, 156}
\definecolor{otherColor}{RGB}{145, 0, 255}
\definecolor{unlabeledColor}{RGB}{0, 0, 0}

\newcommand{\ourmodellong}{Lift, Splat, Map}
\newcommand{\ourmodel}{\textsc{LSMap}}
\newcommand{\coda}{CODa}

\newcommand{\seclabel}[1]{\label{sec:#1}}

\newcommand{\sseclabel}[1]{\label{ssec:#1}}
\newcommand{\ssecref}[1]{Sec.~\ref{ssec:#1}}

\newcommand{\figlabel}[1]{\label{fig:#1}}
\newcommand{\figref}[1]{Fig.~\ref{fig:#1}}
\newcommand{\tablabel}[1]{\label{tab:#1}}
\newcommand{\tabref}[1]{Table~\ref{tab:#1}}
\newcommand{\eqnlabel}[1]{\label{eq:#1}}
\newcommand{\eqnref}[1]{Eq.~\ref{eq:#1}}
\newcommand{\alglabel}[1]{\label{alg:#1}}
\newcommand{\algref}[1]{Alg.~\ref{alg:#1}}

\newcommand{\redacted}{\textbf{REDACTED}}



%% file: introduction.tex
\section{Introduction}
\seclabel{introduction}

Modern robot navigation methods~\cite{yurtsever2020survey, wang2022agriculture, ventura2012search} benefit heavily from context aware scene representations. However, learning scalable scene representations for urban environments is difficult due to heavy occlusions, unstructured environments, and significant class diversity. 

Semantic scene completion (SSC)~\cite{song2017semantic} models produce robust representations for highly occluded and unstructured environments. These models receive image and LiDAR data and predict the semantic class for each voxel in the local scene. SSC predictions are readily usable by planning and navigation algorithms, and are commonly used in a variety of domains~\cite{philion2020lift, meng2023terrainnet, liang2020sscnav}. Furthermore, SSC methods~\cite{meng2023terrainnet, liu2023bevfusion} that simplify the representation to bird's eye view (BEV) can operate in real-time with limited compute.

\begin{figure*}
    \centering
    \includegraphics[width=\textwidth]{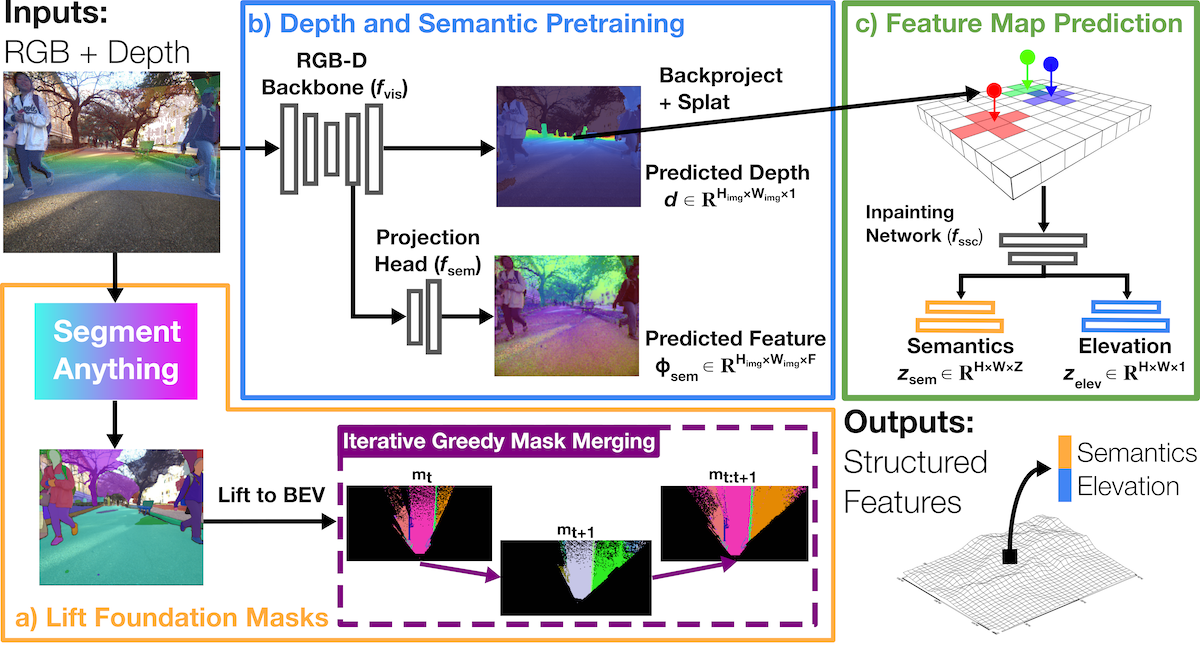}
    \caption{\textbf{\ourmodel{} Architecture and Training Pipeline.} a) We lift instance masks from SegmentAnything~\cite{kirillov2023segment} to bird's eye view (BEV) space. We greedily merge sequential masks based on mask intersection area and use contrastive loss to learn continuous semantic representations using these BEV masks. b) We pretrain a feature and depth completion backbone using Dino~\cite{oquab2023dinov2} features and ground truth depth labels. c) \ourmodel{} predicts and splats semantic features to a BEV feature map and uses a multi-head inpainting network to predict a continuous representation composed of semantic and elevation features. For more details, please see \ssecref{technicalformulation}.}
    \figlabel{mainarchitecture}
    \vspace{-20pt}
\end{figure*}

However, these methods require expensive human annotations and pre-enumerating a list of semantic classes. This limits their scalability to urban environments, where there exist an unbounded number of environments and semantic classes. Furthermore, modern SSC methods predict the semantic classes for terrains and objects jointly, meaning that the terrain underneath objects is unknown. While various approaches~\cite{karnan2023sterling, kavan2021vrlpap, jung2023v} address this by leveraging self-supervised learning to learn terrain aware representations, they are unable to reason about occluded regions and only learn the traversability of terrains. 

Towards overcoming these limitations, we introduce \ourmodel{}, a novel approach to learning context aware representations for complex, dynamic environments. Our pipeline \textbf{lifts} 2D segmentation masks from visual foundation models like SegmentAnything~\cite{kirillov2023segment} and \textbf{splats} these masks to BEV. We merge sequential masks to construct a local \textbf{map}, enabling us to learn an encoder that maps high-dimensional RGBD images to continuous semantic and elevation embeddings for the entire scene. Importantly, our encoder is robust to occlusions, open set, and can be trained without human labels, providing a representation useful for fine-grained downstream tasks, such as semantic scene completion (SSC) or elevation scene completion (ESC).

To evaluate \ourmodel{}, we test our representation on semantic and elevation scene completion tasks. We compare to TerrainNet~\cite{meng2023terrainnet}, a state of the art baseline for supervised SSC and ESC. Furthermore, we compare our representation against directly using Dino~\cite{zhang2022dino} and Dinov2~\cite{oquab2023dinov2} features for unsupervised SSC. We find that \ourmodel{} outperforms existing approaches on both supervised and unsupervised scene completion tasks.

The key contributions of this paper are as follows: 1) \ourmodel{}, a novel approach that learns open set semantic aware representations for downstream tasks without human labels 2) A pipeline for generating high quality BEV segmentation masks and ground truth depth images in dynamic environments 3) Detailed evaluation of \ourmodel{} against baseline methods for both supervised and unsupervised SSC, demonstrating the effectiveness of our feature representation for downstream tasks.

%% file: relatedwork.tex
\section{Related Works}
\seclabel{relatedworks}

In this section, we review existing works that learn semantic aware representations for scene completion. We focus on supervised and self-supervised methods. In this context, we define supervised methods as those requiring human labels.

\subsection{Supervised Methods}
\sseclabel{supervisedmethodsrelatedworks}

Supervised methods for semantic scene understanding have been deployed extensively in robotics domains. We limit the discussion to methods designed for outdoor, real-world deployments.

\textbf{Semantic Scene Completion (SSC):} Outdoor methods are typically evaluated on autonomous driving benchmarks~\cite{behley2019semantickitti, li2023sscbench}, as few benchmarks exist for urban robots. State-of-the-art 3D SSC approaches trade off inference speed for accuracy~\cite{cheng2020s3cnet, zou2021up, rist2021semantic, jiang2023symphonize, li2023voxformer} or accuracy for inference speed~\cite{wilson2022motionsc, roldao2020lmscnet}. Most 3D SSC methods only predict semantic occupancy, leading to BEV methods~\cite{meng2023terrainnet, liu2023bevfusion}, that predict additional values in a unified BEV space, such as separate object, terrain, and elevation maps. While these approaches are promising for urban and off-road environments, they require large amounts of training data and suffer from data and label shift\cite{liu2022undoing, karnan2022voila}.

\begin{wrapfigure}{r}{0.5\textwidth}
    \vspace{-10pt}
    \centering
    \includegraphics[width=0.5\textwidth]{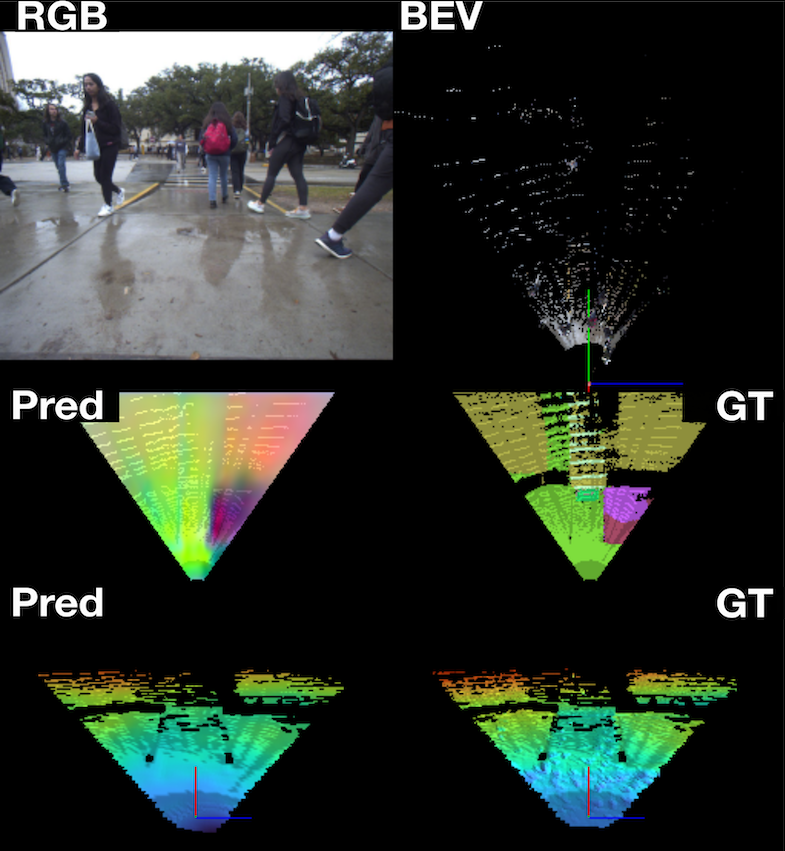}
    \caption{\textbf{\ourmodel{} Predictions.} Our model takes RGB and LiDAR depth measurements in the form of RGB-D images (first row), and predicts continuous semantic features and elevation for the entire field of view. We perform PCA dimensionality reduction on the continuous semantic features for visualization. Patches shaded in white correspond to unoccluded regions.}
    \figlabel{pcasamlabel}
    \vspace{-12pt}
\end{wrapfigure}

\subsection{Self-Supervised Methods}
\sseclabel{selfsupervisedmethodsrelatedworks}

There exist a plethora of methods for semantic scene understanding that learn continuous or hard semantic representations without human labels.

\textbf{Continuous Representation Learning:} Continuous representation learning approaches do not require pre-enumerating the list of classes and learn feature representations using auxiliary task objectives. These auxiliary tasks range from traversability costmap learning~\cite{karnan2023sterling, kavan2021vrlpap, jung2023v, roth2023viplanner, karnan2023wait} to anticipating physical terrain properties~\cite{loquercio2023learning, sathyamoorthy2022terrapn, wellhausen2019should}. These approaches produce continuous embeddings for static, unoccluded scene regions that are well-aligned to the auxiliary task.

\textbf{Hard Representation Learning:} Hard representation learning methods leverage pseudo-labels to supervise representation learning. While this is well-explored for image segmentation~\cite{van2021unsupervised, van2022discovering, cho2021picie, hamilton2022unsupervised}, it is less studied for SSC tasks. S4C~\cite{hayler2023s4c} is a 3D SSC model that leverages off-the-shelf image segmentation networks to supervise the prediction of hard semantic labels. Another approach~\cite{ovsep2024better} leverages visual language models\cite{kirillov2023segment, radford2021learning} to generate instance masks for training a panoptic segmentation network on 3D point clouds. These approaches both require pre-enumerating a list of classes for training and deployment.

\textbf{Hybrid Approach:} Our approach is most similar to S4C~\cite{hayler2023s4c} as we lift class-agnostic segmentation labels from visual foundation models for SSC. In contrast to S4C, we use instance masks from SegmentAnything~\cite{kirillov2023segment}, an instance segmentation model, to learn \textit{continuous} semantic aware feature representations. This enables generalization to an open set of semantic classes. Furthermore, S4C is designed to predict semantic occupancy in pre-traversed environments while our approach generalizes to \textit{novel} environments.

%% file: approach.tex
\section{Approach}
\seclabel{approach}

\subsection{Preliminary}
\textbf{Problem setup.} We aim to predict a dense BEV scene within the current field of view of the robot, given synchronized RGB images and 3D point clouds. We compute the RGBD image $\mathbf{o_t}$ by projecting each 3D point \textit{(x, y, z)} to a 2D pixel \textit{(u, v)} using the camera intrinsics \textbf{K} and extrinsics [\textbf{R, t}] in \eqnref{lidarprojection}. 

\begin{equation}
\begin{bmatrix}
u, v, d
\end{bmatrix}^T =
K
\begin{bmatrix}
I_{3 \times 3} & 0_{3 \times 1}
\end{bmatrix}
\begin{bmatrix}
\mathbf{R} & \mathbf{t} \\
\mathbf{0}_{1 \times 3} & 1
\end{bmatrix} 
\begin{bmatrix} x, y, z, 1 \end{bmatrix}^T
\eqnlabel{lidarprojection}
\end{equation}

Given the current RGBD image $\mathbf{o_t}$, we learn a neural network $\Theta$ that outputs a bird's eye view (BEV) map $\mathbf{Y_t}\in \{z_{\mathrm{sem}}, z_{\mathrm{elev}} \}^{H\times W \times (Z+1)}$ , where H and W denote the height and width of the map. Each map cell contains a semantic feature $z_{\mathrm{sem}}\in \mathbb{R}^Z$ and elevation $z_{\mathrm{elev}}\in \mathbb{R}^1$. The predicted BEV map $\hat{Y}_t = \Theta(o_t)$ must match the ground truth map $Y_t$ as closely as possible. We accomplish this in two stages: \textsc{1) Depth and semantic feature pre-training 2) Representation learning for semantic scene completion}.

\subsection{Technical Formulation}
\sseclabel{technicalformulation}

\textbf{Depth and Semantic Feature Pre-training.} Motivated by recent visual foundation model works~\cite{zhang2022dino, oquab2023dinov2}, we first learn a visual representation $\phi_{\mathrm{vis}}=f_{\mathrm{vis}}(o_t)$ that produces an embedding $\phi_{\mathrm{vis}} \in \mathbb{R}^K$ for each pixel in the image $o_t$. Following prior work~\cite{meng2023terrainnet}, we use a depth completion head $d = f_{\mathrm{depth}}(\phi_{\mathrm{vis}})$, where $d\in \mathbb{R}^{H_{\mathrm{img}} \times W_{\mathrm{img}} \times C}$ and $H_{\mathrm{img}}$, $W_{\mathrm{img}}$, and \textit{C} are the image height, image width, and number of depth bins. We find that representations $\phi_{\mathrm{vis}}$ trained using only depth-based loss become biased towards depth completion rather than semantic segmentation. Thus, we introduce an additional semantic head $\phi_{\mathrm{sem}} = f_{\mathrm{sem}}(\phi_{\mathrm{vis}})$ to distill features from visual foundation models~\cite{zhang2022dino, oquab2023dinov2}. These features are effective for both depth estimation and semantic segmentation.

Prior works~\cite{yang2023emernerf, yang2024denoising} demonstrate that positional embedding artifacts in visual foundation models adversely affect performance. Thus, we enforce $l^2$ consistency for pixel features that project to the same BEV patch across different views. This gives us \eqnref{pretrainloss}, the full pre-training objective with a multi-view consistency ($\mathcal{L}_{\mathrm{mv}}$), foundation model ($\mathcal{L}_{\mathrm{fdn}}$), and depth prediction ($\mathcal{L}_{\mathrm{depth}}$) loss. We refer the reader to Appendix \ssecref{optimization} for an explanation of each loss term.

\begin{equation}
\begin{aligned}
    \mathcal{L}_{\mathrm{pretrain}} &= \alpha_1 \mathcal{L}_{\mathrm{mv}} + \alpha_2 \mathcal{L}_{\mathrm{fdn}} + \alpha_3 \mathcal{L}_{\mathrm{depth}}
\end{aligned} \eqnlabel{pretrainloss}
\end{equation}

\textbf{Learning Continuous Representations for Semantic Scene Completion} 

To learn a continuous representation for SSC, we train a neural network $f_{\mathrm{ssc}}$ to learn the function $\mathbf{Y_t} = f_{ssc}(S_t)$, where $S_t$ is a set of 3D point features $\{(x, y, z, \phi) \}$. To construct $\phi$, we first apply a multi-layer perceptron (MLP) on \textit{z} to obtain an elevation embedding $\phi_{\mathrm{elevation}} = MLP(z)$, concatenate the visual and elevation embeddings, and apply another MLP as follows: $\phi=MLP([\phi_{\mathrm{vis}}, \phi_{\mathrm{elevation}}])$. We follow prior work~\cite{meng2023terrainnet} and use a \textit{soft quantization} approach to splat each point feature $\phi$ to the neighboring BEV map cells, which we depict using the \textit{splat} operation in \eqnref{fullmodeleqn}. We use a U-Net~\cite{ronneberger2015unet} multi-head inpainting module with a shared encoder and separate decoder heads for predicting the semantic feature $Z_{\mathrm{sem}}=\{z_{\mathrm{sem, i}} | \, i \in I \}$ and elevation maps $Z_{\mathrm{elev}}=\{z_{\mathrm{elev, i}} | \,i \in I \}$, where $I \equiv \{\mathrm{1,...,HW}\}$ is the set of indices for all patches in the current feature map $Y_t$.

\begin{equation}
\begin{aligned}
Y_t = [Z_\mathrm{sem}, Z_\mathrm{elev}] &= f_\mathrm{ssc}(\textrm{splat}(\{x, y, z, \phi \}) )
\end{aligned} \eqnlabel{fullmodeleqn}
\end{equation}

To supervise our training, we require a BEV segmentation label $m_t\in \mathbb{Z}^{H \times W \times 1}$, which we describe in \ssecref{maskdatasetdetails}. We use a contrastive loss to attract embeddings in $Z_{\mathrm{sem}}$ with the same label and repel those that are different. Specifically, we use supervised contrastive loss~\cite{kohsla2020supcon}, which demonstrates strong results when compared to cross-entropy or self-supervised contrastive losses. To define our objective, let $P(i)\, \equiv \, \{ p \in A(i) : \, m(p) = m(i) \}$ be the set of indices for all patches with the same mask label as patch \textit{i} and $A(i)\, \equiv \, I \setminus \{i\}$ be the set of indices for all patches excluding the current patch \textit{i}. We define the SSC training objective as follows, where $\beta_1$, $\beta_2$, and $\beta_3$ are tunable hyperparameters.

\begin{equation}
\begin{aligned}
\mathcal{L}_{\mathrm{supcon}} &= \frac{1}{|I|} \sum_{i \in I} \frac{-1}{|P(i)|} \sum_{p \in P(i)} \log \frac{\exp (z_{\mathrm{sem,} i} \cdot z_{\mathrm{sem,} p} / \tau)}{\sum_{a \in A(i)} \exp (z_{\mathrm{sem,} i} \cdot z_{\mathrm{sem,} a} / \tau)} \\
\mathcal{L}_{\mathrm{elev}} &= \frac{1}{|I|} \sum_{i \in I} |\hat{z}_{\mathrm{elev,} i} - z_{\mathrm{elev,} i}|^{1}_{1}\\
\mathcal{L} &= \beta_1 \mathcal{L}_{\mathrm{supcon}} + \beta_2 \mathcal{L}_{\mathrm{elev}} + \beta_3 \mathcal{L}_{\mathrm{depth}}
\end{aligned}
\label{eq:lossformulation}
\end{equation}

%% file: implementation.tex
\section{Implementation Details}
\seclabel{implementation}

In this section, we describe our procedure for generating pseudo ground truth depth and BEV segmentation labels. Our method requires these labels for the technical formulation in \ssecref{technicalformulation}. We conclude this section with a discussion of ground truth BEV label generation procedure and model architecture.

\subsection{Hardware Setup}
\sseclabel{hardwaresetup}

We use the semantic segmentation split of \coda{}~\cite{zhang2023towards} to construct the ground truth SSC and ESC datasets. During training, \ourmodel{} requires synchronized stereo RGB and 3D LiDAR sensor data, known camera intrinsic and extrinsic calibrations, and locally consistent robot poses. During deployment, we only require a single synchronized RGB camera, 3D LiDAR, and camera intrinsic and extrinsics.

\subsection{Ground Truth Depth Dataset}
\sseclabel{depthdatasetdetails}

Prior depth completion datasets~\cite{uhrig2017sparsity} demonstrate that naively projecting single scan LiDAR yields sparse depth labels that degrade depth completion network performance. Other work~\cite{meng2023terrainnet} alleviates this problem by accumulating LiDAR point clouds across sequential observations before projection. This method fails in urban environments due to dynamic agents, which leave streaking artifacts or sparse depth labels for dynamic pixels as seen in \figref{qualitativedepthanalysis}.

For our dataset, we combine the methodology of the KITTI depth evaluation suite~\cite{uhrig2017sparsity} and inverse distance weighting (IDW) for depth infilling~\cite{premebida2016high}. For each timestamp t, we accumulate and project the previous 50 LiDAR point clouds to the current image. We compute the stereo depth and filter LiDAR depth measurements with large discrepancies and apply edge-aware kernel filter to upsample the depth measurements. We find this step is key to filtering noisy depth labels resulting from parallax, sensor measurement differences, and dynamic objects. For more information and qualitative results, please refer to Appendix~\ssecref{qualitativedepthanalysis}.

\subsection{Lifting Segmentation Masks to Bird's Eye View} 
\sseclabel{maskdatasetdetails}

Recent zero-shot segmentation models like SegmentAnything~\cite{kirillov2023segment} demonstrate a strong ability to group semantically similar image pixels. We leverage this attribute to precompute a label mask $m_{\mathrm{img,} t} = \mathrm{segment}(o_t)$, defining $m_{\mathrm{img,} t} \in \mathbb{Z}^{H_{\mathrm{img}} \times W_{\mathrm{img}} \times 1}$. We backproject each pixel in $m_{\mathrm{img,} t}$ using \eqnref{pixelprojection} to obtain a point feature set $\{(x, y, z \} \sqcup m_{\mathrm{img,} t}) \}$ and project down to a BEV map using the nearest integer grid coordinates. Taking the argmax of the mask label frequencies for each cell in the local map gives the final BEV map $m_t \in \mathbb{Z}^{H \times W \times 1}$. We repeat this procedure for multiple observations of the same scene to ensure all occluded regions are labeled.

\begin{equation}
\begin{bmatrix}
x, y, z, 1
\end{bmatrix}^T =
\begin{bmatrix}
\mathbf{R^T} & \mathbf{-R^T t} \\
\mathbf{0}_{1 \times 3} & 1
\end{bmatrix} 
\begin{bmatrix}
I_{3 \times 3} \\
0_{1 \times 3}
\end{bmatrix}
K^{-1}
\begin{bmatrix} u, v, d \end{bmatrix}^T
\eqnlabel{pixelprojection}
\end{equation}

While class-agnostic segmentation models enable learning open-set representations, they create label inconsistencies between classes across different masks. To alleviate this, we propose an Iterative Greedy Mask Merging (IGMM) strategy. Given a pair of overlapping masks $m_1$ and $m_2$, IGMM computes a one-to-one map from labels in $m_2$ to $m_1$ by greedily assigning labels with the highest intersection across masks. All unassigned labels in $m_2$ are merged as new labels in the accumulated mask. We formalize our algorithm in Appendix~\ssecref{igmm}, \algref{igmm} and present qualitative results of the lifted segmentation masks in Appendix~\ssecref{igmm}, \figref{sammergefigure}.

It is important to only project mask labels corresponding to static entities, as dynamic masks introduce multiview inconsistencies to the merged map. Thus, we construct an image mask $\mathrm{mv} \in \{0, 1\}^{H_{\mathrm{img}} \times W_{\mathrm{img}}}$ where $\{0, 1\}$ represents whether or not the pixel corresponds to a movable entity. We obtain $\mathrm{mv}$ by constructing a static point cloud map $M\in \mathbb{R}^{N \times 3}$ using the Voxblox~\cite{oleynikova2017voxblox} library and multiview point cloud observations of the same scene. Then, we perform a K-nearest neighbors occupancy check for each point in the query point cloud using the static map. Points that do not contain at least K neighboring points within a ball distance m are considered dynamic. We project the dynamic points to the image using \eqnref{lidarprojection} to obtain $\mathrm{mv}$.

\subsection{Generating Ground Truth Semantic and Elevation Maps}
We closely follow Meng et al.~\cite{meng2023terrainnet} to construct a ground truth BEV semantic map by accumulating labeled 3D point clouds using known robot poses. More specifically, we accumulate 50 point clouds for each BEV frame and assign the most frequent class label for each map cell. To construct the elevation map, we average the bottom 3 static points for each map cell. For all maps, we use a map size of 25.6m $\times$ 25.6m with a resolution of 0.1 m. We limit the minimum and maximum elevation range to -1.2m and 1.8m respectively. 

\subsection{Model Architecture Discussion}
\sseclabel{modelarchitecturedetails}

We closely follow the model architecture of our supervised model baseline, TerrainNet~\cite{meng2023terrainnet}, achieving similar real-time inference speed. Additionally, this reduces the confounding effects of model architectural differences, allowing us to fairly compare the learned representations. We refer readers to the original work for more information and describe any architectural modifications here. 

We append an additional convolutional projection head to the penultimate layer of the RGB-D backbone to encourage learning features that are both semantically meaningful and useful for depth prediction. We use this projection head during pre-training and drop it when training the full model. We modify the semantic decoder head of the inpainting network to produce a semantic representation with 64 dimensions rather than the number of semantic classes.

%% file: experiments.tex
\section{Experiments}
\label{sec:experiments}

In this section, we describe the experiments performed to evaluate \ourmodel{}. We tailor our experiments to answer the following questions:
\begin{itemize}[leftmargin=0.15in]
    \item (\textit{$Q_1$}) How effective are \ourmodel{} features for improving task-specific performance in comparison to existing baselines on supervised semantic and elevation scene completion?
    \item (\textit{$Q_2$}) How effective are \ourmodel{} features in comparison to visual foundation model features for unsupervised semantic scene completion? 
\end{itemize}

\textbf{Dataset.} We evaluate both experiments on \coda{}~\cite{zhang2023towards}, a large-scale, real-world urban robot dataset collected on a university campus. We pretrain \ourmodel{} on 70\% of the total frames in \coda{} and ensure these frames are excluded from the annotated validation and test splits. For supervised training and evaluation, we use the semantic segmentation split of \coda{}, randomly selecting 70\%, 15\%, and 15\% of the frames for training, validation, and testing. In total, \coda{} contains over 5000 annotated 3D point clouds with 23 terrain semantic classes from indoor, road, and hybrid traffic scenes under rainy, dark, sunny, and cloudy weather conditions.

\textbf{Evaluation Metrics.} To evaluate the SSC task, we use the standard Intersection over Union (IoU) metric. Notably, we evaluate the IoU separately for occluded and unoccluded regions in addition to both regions combined. This provides insight into how robust various methods are to occlusions. For the ESC task, we report the Mean Absolute Error (MAE) in terms of meters. For \textit{$Q_1$}'s experiments, we train on the semantic segmentation train split provided by \coda{} and report results on the test split. For $Q_2$, we follow the standard evaluation procedure for unsupervised semantic segmentation~\cite{hamilton2022unsupervised}: 1) Cluster features from the annotated validation split using Kmeans 2) Map cluster centroids to labels using the Hungarian algorithm on the test split 3) Compute IoU on the annotated test split using the pre-trained K-nearest neighbors and cluster to label mapping.

\subsection{Supervised Scene Completion Evaluation}
\sseclabel{supervisedscenecompletionevaluation}

\input{tables/sup_ssc_tb}

It is common practice to pre-train a feature representation without labels and finetune this representation for domain-specific tasks. As such, we conduct the following experiments to evaluate the effectiveness of \ourmodel{} representations on scene completion tasks. We evaluate the following approaches for these experiments.

\textbf{Interpolation (GT, Pred).} We provide the ground truth (GT) or predicted (Pred) semantic and elevation map constructed from a single frame as input to the model. We bilinearly interpolate the elevation map and perform nearest neighbor label assignment for all occluded cells using the known values. We use TerrainNet~\cite{meng2023terrainnet} and mask occluded regions to obtain unoccluded map predictions.


\textbf{TerrainNet (Baseline).} We train the TerrainNet model from scratch for 50 epochs or until convergence using ground truth labels. This model architecture most closely resembles ours and represents the expected performance without any pre-training. We use the same map size and resolution parameters as \ourmodel{} to minimize model-specific differences.

\textbf{\ourmodel{}+FT.} For this approach, we pretrain \ourmodel{} using pseudo-ground truth SAM labels and elevation maps for 75 epochs or until convergence. Then, we follow the methodology of prior work on transfer learning for semantic segmentation~\cite{henaff2021efficient} and freeze the entire model, replacing the penultimate and final layers of the semantic and elevation decoder heads, and finetuning for 50 epochs or until convergence. This assesses how effective the label-free representation is for downstream tasks.

\textbf{\ourmodel{}+FT+E2E.} This baseline repeats the training process described for \ourmodel{}+FT. After finetuning the decoder heads, we unfreeze the full model and finetune end to end for 50 epochs or until convergence using a 10x lower learning rate than the initial finetuning step. This allows our model to perform minor adjustments to the feature representation for domain-specific tasks. 

\textbf{Analysis.} We find that the \ourmodel{} + FT + E2E representation greatly improves downstream performance on semantic and elevation scene completion for all metrics. Notably, \tabref{sup_ssc_tb} shows finetuning \ourmodel{}'s representation improves SSC performance by at least 7.5\% compared to all other approaches. Furthermore, while \ourmodel{} + FT does not exceed other baselines for the SSC task, it falls behind our TerrainNet baseline by only 5 percent for unoccluded regions, demonstrating that our representation can learn low-level features comparable to those learned by a model trained from scratch with labels. Interestingly, we find that the TerrainNet baseline performs better on occluded regions than unoccluded. We believe this is due to our model learning to overpredict less frequent classes to improve overall performance. We do not observe this behavior in our model, suggesting that our model's representation leverages the scene context more effectively.

\subsection{Unsupervised Semantic Scene Completion Evaluation}
 
\input{tables/unsup_ssc_tb}

For this experiment, we evaluate how well our feature representation can discriminate between semantic classes without access to ground truth labels. We train \ourmodel{} using pseudo-ground truth segmentation labels for 75 epochs and compare our approach against naively using foundation model features from Dino~\cite{zhang2022dino} and Dinov2~\cite{oquab2023dinov2}. We now describe our process for building feature maps from foundation model features.

\textbf{Building Foundation Model Feature Maps.} We first compute a dense feature map using feature extractors from visual foundation models. For computational reasons, we follow existing works that distill foundation model features~\cite{yang2023emernerf} and reduce the feature map dimension to 64 using Principal Component Analysis (PCA). Using \eqnref{pixelprojection}, we backproject image features to a local BEV map, aggregating features for each map cell via max pooling. We find this aggregation method gives the best results, aligning with findings from other works~\cite{keetha2023anyloc}. This gives us our distilled feature map $Y^{fdn}_t \in \{z_{fdn}\}^{H \times W \times 64}$ that we use for evaluation. It is important to note that we use the same feature dimension for \ourmodel{} embeddings.

\textbf{Analysis.} \tabref{unsup_ssc_tb} shows the per class IoU and mIoU using various pre-trained visual representations. We find that our \ourmodel{} representation outperforms all other methods by at least 6 percent for mIoU. Furthermore, our approach outperforms all other approaches in per class IoU for 10 out of 17 semantic classes. This demonstrates that while foundation model features can generalize to scene completion tasks, encoders trained using our approach yield more semantically discriminative representations for SSC.

%% file: tables/sup_ssc_tb.tex
\begin{table}[htbp]
\vspace{-16pt}
\centering
\begin{tabular}{|c|c|c|c|c|c|c|}
    \hline
    \multirow{2}{*}{Model Type} & \multicolumn{3}{c|}{Semantic} & \multicolumn{3}{c|}{Elevation} \\ \cline{2-7} 
                                & Occluded & Unoccluded & Both & Occluded & Unoccluded & Both \\ \hline
    Interpolation (GT) & 71.2 & 92.8 & 75.2 & 0.080 & 0.055 & 0.075 \\ \hline
    Interpolation (Pred) & 69.1 & 74.6 & 70.5 & 0.087 & 0.046 & 0.079 \\ \hline
    TerrainNet~\cite{meng2023terrainnet} (Baseline) & 80.2 & 74.6 & 79.0 & 0.047 & 0.046 & 0.047 \\ \hline
    \ourmodel{}+ FT (Ours) & 66.6 & 69.6 & 67.4 & 0.060 & 0.045  & 0.057 \\ \hline
    \ourmodel{} + FT + E2E (Ours) & \textbf{85.7} & \textbf{89.2} & \textbf{86.5} & \textbf{0.040} & \textbf{0.034} & \textbf{0.039} \\ \hline
\end{tabular}
\vspace{1em}
\caption{Mean Intersection over Union (mIoU) of different models trained on \coda{}. We report the mIoU in unoccluded, occluded, and both regions as indicated by the respective columns. We describe the baselines in detail in \ssecref{supervisedscenecompletionevaluation}. Bold numbers indicate the highest-performing method for each category.}
\tablabel{sup_ssc_tb}
\vspace{-20pt}
\end{table}

%% file: tables/unsup_ssc_tb.tex

\begin{table*}[ht]
\vspace{-10pt}
\footnotesize
\centering
\setlength{\tabcolsep}{1pt}
\renewcommand{\arraystretch}{1.25} 


\begin{tabular}{cc|c|c|c|c|c|c|c|c|c|c|c|c|c|c|c|c|c|c|}
    \cellcolor{white}{\color{black}\textbf{Approach}} &
    \rotatebox{90}{\cellcolor{white}{\color{black}\textbf{mIoU}}} &
    \rotatebox{90}{\cellcolor{concreteColor}{\color{white}\textbf{Con.}}} &
    \rotatebox{90}{\cellcolor{grassColor}{\color{white}\textbf{Gra.}}} &
    \rotatebox{90}{\cellcolor{rocksColor}{\color{white}\textbf{Roc.}}} &
    \rotatebox{90}{\cellcolor{speedwayBricksColor}{\color{white}\textbf{Spe.}}} &
    \rotatebox{90}{\cellcolor{redBricksColor}{\color{white}\textbf{Red.}}} &
    \rotatebox{90}{\cellcolor{pebblePavementColor}{\color{white}\textbf{Peb.}}} &
    \rotatebox{90}{\cellcolor{lightMarbleTilingColor}{\color{white}\textbf{Til.}}} &
    \rotatebox{90}{\cellcolor{dirtPathsColor}{\color{white}\textbf{Dir.}}} &
    \rotatebox{90}{\cellcolor{roadPavementColor}{\color{white}\textbf{Roa.}}} &
    \rotatebox{90}{\cellcolor{shortVegetationColor}{\color{white}\textbf{Sho.}}} &
    \rotatebox{90}{\cellcolor{metalGratesColor}{\color{white}\textbf{Met.}}} &
    \rotatebox{90}{\cellcolor{woodPanelColor}{\color{white}\textbf{Woo.}}} &
    \rotatebox{90}{\cellcolor{carpetColor}{\color{white}\textbf{Car.}}} &
    \rotatebox{90}{\cellcolor{crosswalkColor}{\color{white}\textbf{Cro.}}} &
    \rotatebox{90}{\cellcolor{domeMatColor}{\color{white}\textbf{Mat}}} &
    \rotatebox{90}{\cellcolor{stairsColor}{\color{white}\textbf{Sta.}}} &
    \rotatebox{90}{\cellcolor{metalFloorColor}{\color{white}\textbf{Oth.}}}\\
    \hline
    Dinov1~\cite{zhang2022dino} & 18.07 & 6.9 & 32.8 & 62.3 & 6.0 & 23.5 & 10.9 & 7.6 & \textbf{17.1} & \textbf{9.0} & 13.6 & \textbf{33.4} & \textbf{25.9} & 0.0 & 14.4 & 4.4 & \textbf{16.8} & 22.0\\
    Dinov2~\cite{oquab2023dinov2} & 11.6 & 4.9 & 21.6 & 26.0 & 9.2 & \textbf{28.6} & 15.2 & 1.9 & 4.7 & 2.6 & 6.7 & 22.2 & 10.7 & 0.0 & 18.8 & 9.6 & 11.7 & 3.5 \\
    \ourmodel{} (Ours) & \textbf{24.1} & \textbf{7.7} & \textbf{48.1} & \textbf{80.3} & \textbf{12.7} & 26.3 & \textbf{21.8} & \textbf{17.2} & 17.0 & 6.6 & \textbf{18.2} & 12.0 & 25.0 & 0.0 & \textbf{33.7} & \textbf{10.8} & 13.9 & \textbf{58.5} \\ 
\end{tabular} 

\caption{Unsupervised SSC evaluation using Dinov1 and Dinov2 foundation model features versus \ourmodel{}. We report mean intersection over union (IoU) and per class IoU. We present the abbreviated class name and describe the full name in Appendix~\ssecref{semanticlabelontology}, \figref{semanticlabelontology}. Bold numbers indicate the highest-performing method for each category.} 

\tablabel{unsup_ssc_tb}
\vspace{-10pt}
\end{table*}

%% file: limitations.tex
\section{Limitations and Future Work}
\seclabel{limitations}

Our approach leverages semantic priors from instance masks predicted by visual foundation models. Thus, we assume that each instance mask corresponds to a unique semantic class, which may not hold for all environments. This problem can be alleviated by utilizing the correlation volume~\cite{teed2020raft} from visual foundation model features to unify disjoint instance masks corresponding to the same semantic class. Extending \ourmodel{} to predict object centric semantic features is a another promising direction for future work and will further close the gap towards scalable context aware scene understanding.

%% file: conclusion.tex
\section{Conclusion}
\seclabel{conclusion}

In this work, we introduce \ourmodellong{}: Lifting Foundation Masks for
Label-Free Semantic Scene Completion (\ourmodel{}), a novel approach to learning continuous representations for context aware scene completion. We lift and merge instance masks from visual foundation models across sequential observations, splat instance masks to a BEV representation, and generate occlusion free BEV segmentation maps. We utilize contrastive learning to learn an encoder that predicts continuous embeddings for the entire BEV scene, including occluded regions. We evaluate \ourmodel{} against similar supervised models on a large scale urban robot dataset and demonstrate that our features transfer well to SSC with lightweight finetuning and greatly outperforms models trained from scratch. We introduce a novel unsupervised semantic scene completion benchmark and find that our representation outperforms standard Dino and Dinov2 features without labels. As the first work to evaluate SSC for robots in urban environments, we release the scene completion datasets in this work to spur future progress towards scalable context aware representation learning.

%% file: appendix.tex
\section{Appendix}
\seclabel{appendix}
\subsection{Semantic Class Ontology}
\sseclabel{semanticlabelontology}
We evaluate \ourmodel{} on 18 classes from the semantic segmentation split of \coda{}~\cite{zhang2023towards}. To balance the class frequencies, we group the original 23 semantic classes into the following labels. Readers can find a detailed summary of this grouping at \redacted{}.

\begin{figure*}[h]
    \vspace{-5pt}
    \centering
    \includegraphics[width=\textwidth]{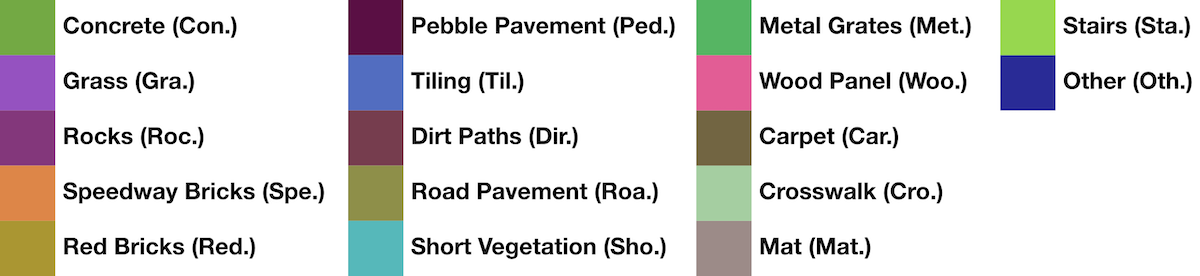}
    \caption{\textbf{Semantic Class Ontology.} We abbreviate the following semantic classes in \tabref{unsup_ssc_tb} for conciseness. Each patch on the left denotes the assigned color label for each class. We follow this color map for the visualizations in this work.}
    \figlabel{semanticlabelontology}
    \vspace{-15pt}
\end{figure*}

\subsection{Qualitative Analysis of Ground Truth Depth}
\sseclabel{qualitativedepthanalysis}

\begin{figure*}[h]
    \vspace{-5pt}
    \centering
    \includegraphics[width=\textwidth]{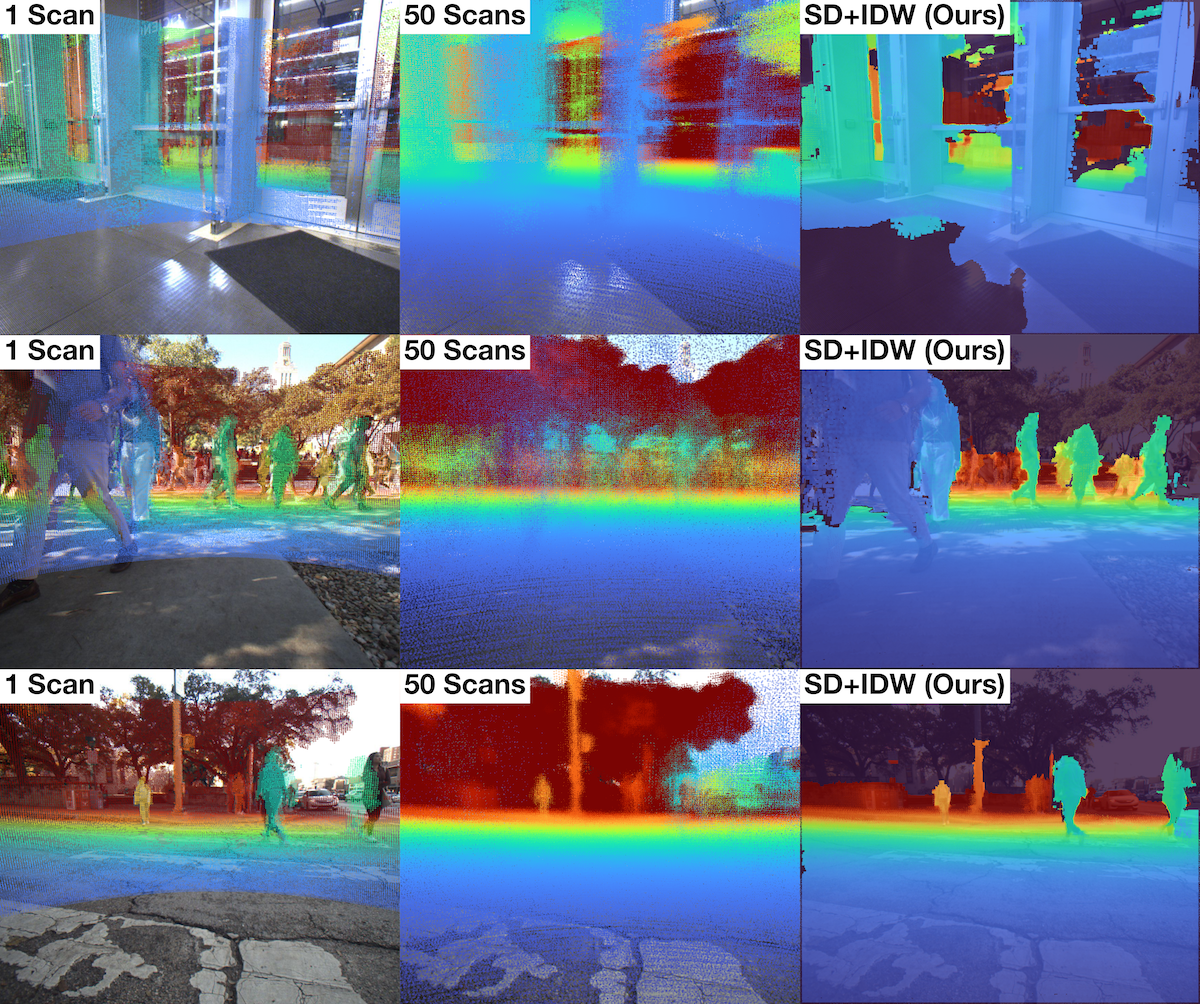}
    \caption{\textbf{Comparison of Different Ground Truth Depths.} We present depth images from three scenes constructed using various depth estimation strategies. We overlay the colorized depth map on the RGB image, where the color indicates the distance from the camera. The left column projects a single LiDAR scan onto the image. The middle column accumulates and projects the past 50 LiDAR scans to the image. The rightmost column uses our proposed depth estimation strategy, Stereo Depth Filtered LiDAR + Inverse Distance Weighting (SD+IDW).}
    \figlabel{qualitativedepthanalysis}
    \vspace{-10pt}
\end{figure*}

In this section, we fully present our depth dataset construction procedure, Stereo Depth Filtered LiDAR + Inverse Distance Weighting (SD+IDW), and qualitative analysis of our ground truth depth labels. 

\textbf{Depth Estimation Procedure.} As mentioned in~\ssecref{depthdatasetdetails}, we combine the methodology of the KITTI depth evaluation suite~\cite{uhrig2017sparsity} and inverse distance weighted (IDW) filtering for depth infilling~\cite{premebida2016high}. For each timestamp t, we perform stereo depth estimation using a semi-global stereo matching algorithm~\cite{hirschmuller2005accurate} to obtain $d^{stereo}_t$ and accumulate 50 sequential LiDAR point clouds before projecting to obtain $d^{accum}_t$. 

We filter LiDAR depth measurements that exhibit large errors relative to the stereo depth measurement. In practice, we find an error threshold of 30\% yields a good balance between filtering erroneous values without over-filtering correct measurements. Finally, we use an IDW kernel filter to upsample the filtered depth measurement for depth completion supervision. We tune the kernel window size to improve depth label density without introducing significant errors in depth.

\textbf{Analysis.} The final filtering step improves depth label density from 30\% to 90\% on average, which prior work~\cite{uhrig2017sparsity} demonstrates is important for non-sparsity invariant CNN architectures. Furthermore, Premebida et al~\cite{premebida2016high} demonstrate that IDW filtering introduces an average error of fewer than 0.06 meters on the KITTI depth dataset~\cite{uhrig2017sparsity}. \figref{qualitativedepthanalysis} presents scenes with sensor synchronization artifacts during heavy rotation (top), streaking artifacts from dynamic agents (middle), and inconsistent depth measurements due to temporary occlusions from dynamic agents (bottom). Our method is less susceptible to erroneous depth measurements caused by these conditions compared to existing approaches. Furthermore, SD+IDW generates far denser depth labels, which further improves model performance.

\subsection{Qualitative Analysis of Continuous Semantic Representation}

\begin{figure*}[htbp]
    \vspace{-25pt}
    \centering
    \includegraphics[width=0.9\textwidth]{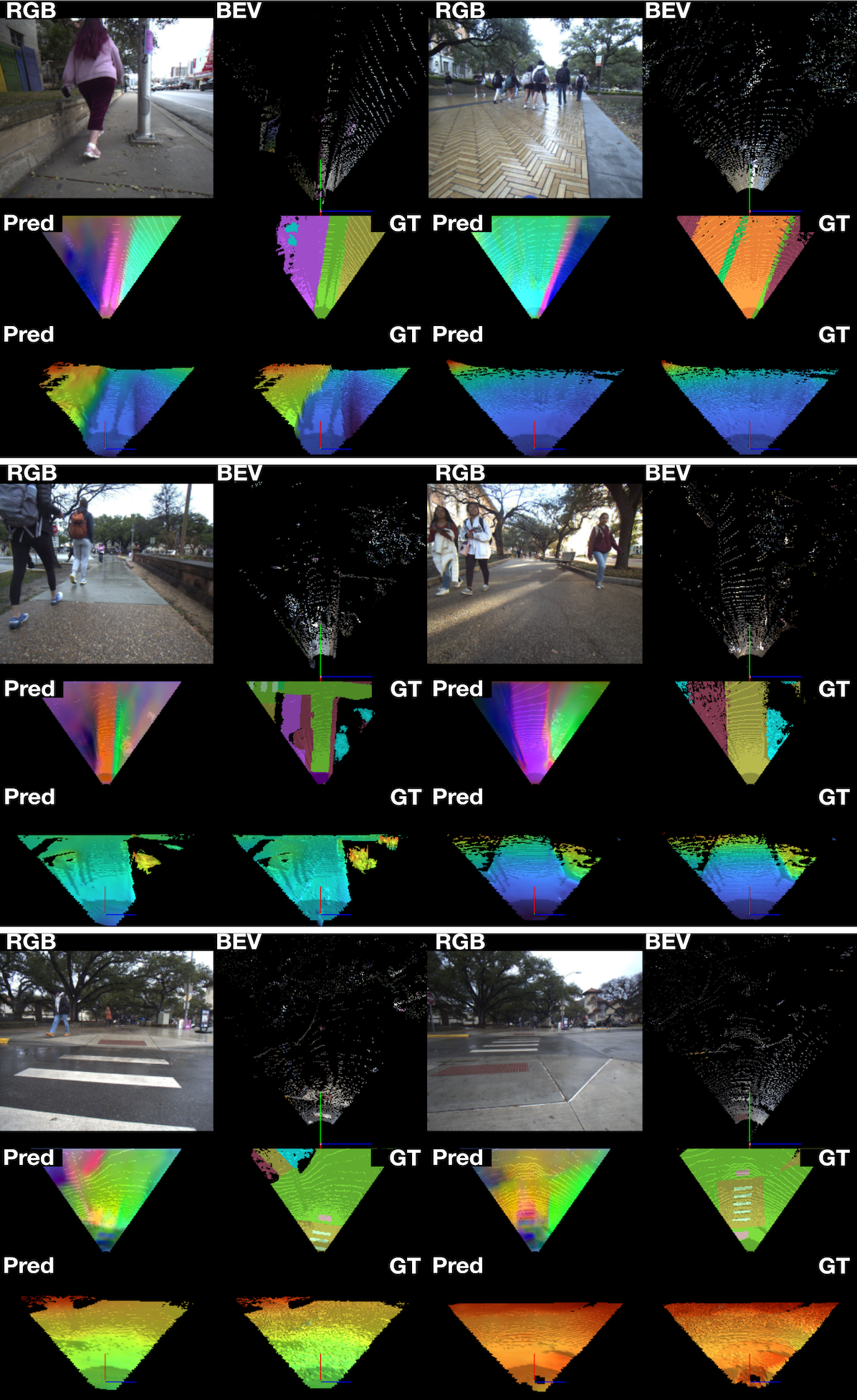}
    \caption{\textbf{Visualization of the Predicted BEV Representations From \ourmodel{}.} We visualize the continuous representation predicted by \ourmodel{} for various scenes in \coda{}. We perform PCA dimensionality reduction to visualize the predicted semantic embeddings for each frame in RGB. For each scene, we show the RGB and LiDAR depth input to our model and predicted semantic and elevation embeddings on the left column. The right column shows the ground truth semantic class (top) and elevation (bottom). Notably, our model learns the following representation without human labels. Patches shaded in white correspond to unoccluded regions.}
    \figlabel{qualitativemodelanalysis}
    \vspace{-5pt}
\end{figure*}

In this section, we qualitatively analyze \ourmodel{} embeddings from diverse scenes in the test set of \coda{}~\cite{zhang2023towards}. \figref{qualitativemodelanalysis} presents comparisons of the predicted continuous semantic representation against the ground truth semantic and elevation maps. It is important to note that \ourmodel{}'s representation is trained completely without human labels. Furthermore, our model predicts embeddings for all BEV patches in the scene, including those in occluded regions and underneath dynamic objects.

\textbf{Analysis.} We find that \ourmodel{} can learn a continuous semantic representation that roughly captures the semantic details. In the top left example, our model can accurately distinguish between the road, sidewalk, and grass regions. In the second row examples, our model can correctly infill features in occluded regions. In the third row examples, our model can distinguish fine-grained details, such as individual crosswalk lines and small sidewalk walkways. 

\subsection{Iterative Greedy Masking Merging Algorithm (IGMM)}
\sseclabel{igmm}

\begin{figure*}[htbp]
    \vspace{-5pt}
    \centering
    \includegraphics[width=\textwidth]{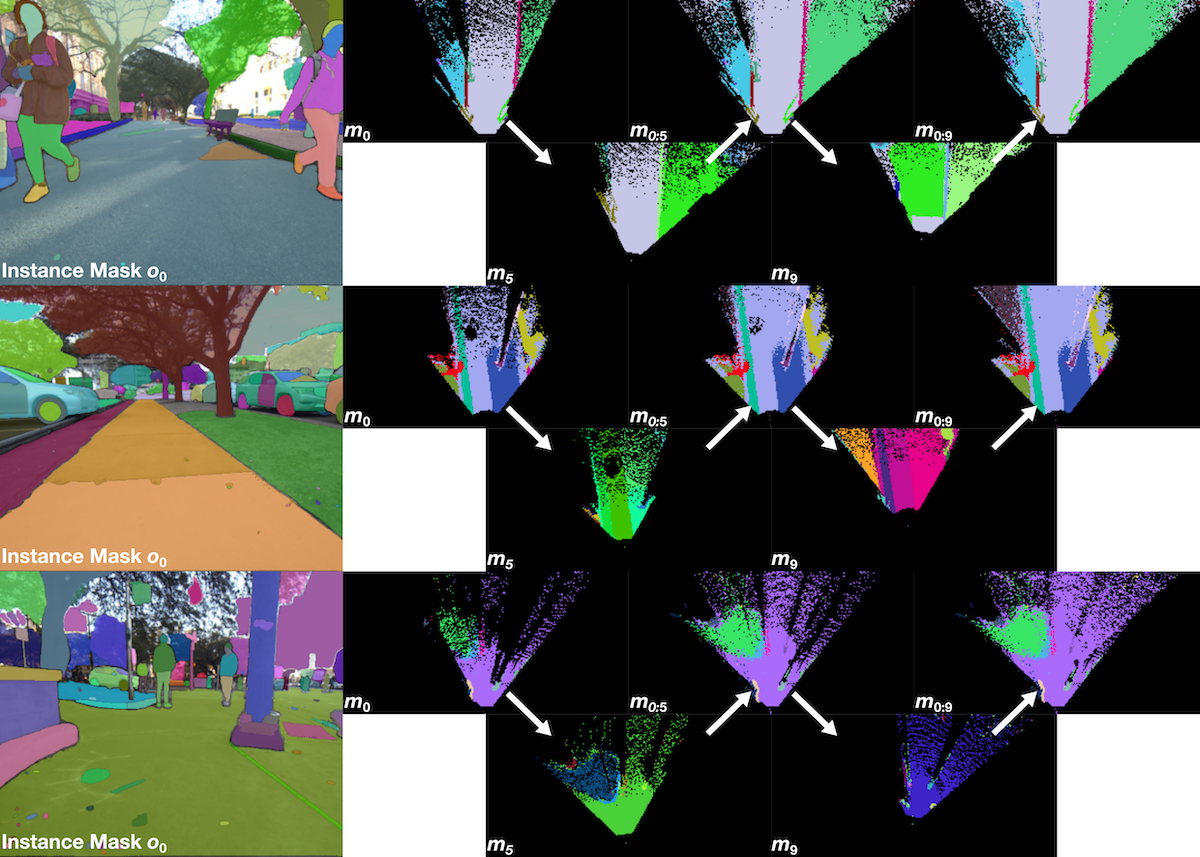}
    \caption{\textbf{Visualizations of IGMM Generated BEV Mask Labels.} We visually present the mask label construction procedure using Iterative Greedy Mask Merging (IGMM) for three scenes in the train split of \coda{}. We overlay SegmentAnything~\cite{song2017semantic} mask labels on the RGB input image and project the labels to BEV space. The top row of each scene shows the merged mask $m_{t_1:t_2}$ from timestamps $t_1$ to $t_2$. The bottom row shows the single-frame mask label. IGMM can merge mask labels from multiple frames to construct an accurate BEV mask despite sensing occlusions and dynamic agents.
    } 
    \figlabel{sammergefigure}
    \vspace{-10pt}
\end{figure*}

Because back projecting individual views of a scene may contain unobserved regions in BEV space, we iteratively merge instance masks from sequential RGD observations $o_{1:T}$. We take the first observation $o_1$ and backproject its instance mask to BEV space to compute a BEV mask $m_1$. For each successive view, we repeat this procedure and use IGMM to merge $m_t$ with $m_1$. 

The full IGMM algorithm presented in \algref{igmm} computes a one-to-one mapping from the labels in $m_t$ to $m_1$. For each unique label in $m_t$, it computes the number of patch overlaps with the labels in $m_1$. It assigns each unique label to the corresponding label in $m_1$ with the most overlaps, skipping labels without any nonzero overlaps. Finally, for all unassigned labels in $m_t$, it assigns new labels sequentially starting from the last label index in $m_1$. 

\begin{algorithm}
\caption{Iterative Greedy Mask Merging (IGMM)}
\begin{algorithmic}[1]
\REQUIRE Masks \(m_1\) and \(m_2\) of size \(H \times W \times 1\)
\STATE Initialize identity label map for \(m_1\)
\STATE \( \text{max\_label} \gets \max(m_1) \)
\STATE \( \text{label\_map} \gets \text{empty map} \)

\FOR{each unique label \( \ell_2 \) in \( m_2 \)}
    \STATE Find the label \( \ell_1 \) in \( m_1 \) with the most overlapping pixels with \( \ell_2 \)
    \STATE \( \text{overlap\_count} \gets \{ \ell_1 : |\{ (i,j) \mid m_1(i,j) = \ell_1 \text{ and } m_2(i,j) = \ell_2 \}|\} \)
    \IF{ \(\text{overlap\_count} \) is not empty}
        \STATE \( \text{max\_overlap\_label} \gets \arg\max \text{overlap\_count} \)
        \STATE Map \( \ell_2 \) to \( \text{max\_overlap\_label} \) in \( \text{label\_map} \)
    \ELSE
        \STATE \( \text{max\_label} \gets \text{max\_label} + 1 \)
        \STATE Map \( \ell_2 \) to \( \text{max\_label} \) in \( \text{label\_map} \)
    \ENDIF
\ENDFOR

\FOR{each pixel \((i,j)\) in \( m_2 \)}
    \IF{ \( m_2(i,j) \neq 0 \) }
        \STATE \( m_2(i,j) \gets \text{label\_map}[m_2(i,j)] \)
    \ENDIF
\ENDFOR

\end{algorithmic}\alglabel{igmm}
\end{algorithm}

\subsection{Optimization}
\sseclabel{optimization}

\textbf{Loss functions.} As discussed in \ssecref{technicalformulation}, our pre-training $\mathcal{L}_{\mathrm{pretrain}}$ and end-to-end training $\mathcal{L}$ losses are as follows:

\begin{equation}
\begin{aligned}
\mathcal{L}_{\mathrm{pretrain}} &= \alpha_1 \mathcal{L}_{\mathrm{mv}} + \alpha_2 \mathcal{L}_{\mathrm{fdn}} + \alpha_3 \mathcal{L}_{\mathrm{depth}}\\
\mathcal{L} &= \beta_1 \mathcal{L}_{\mathrm{contrast}} + \beta_2 \mathcal{L}_{\mathrm{contrast}} + \beta_3 \mathcal{L}_{\mathrm{depth}}
\end{aligned}
\end{equation}

\textbf{Definitions.} Let $V(i)\, \equiv \, \{ v \in I : \, \mathrm{patch}(v) = \mathrm{patch}(i) \}$ be the set of indices for all pixels in image V that backproject to the same BEV patch as pixel i from the anchor image. Let $I \equiv \{1,...,H_{\mathrm{img}}W_{\mathrm{img}} \}$ be the set of indices for all image pixels. Let $I^{\mathrm{anchor}} \equiv \{ i \in I : |V(i)| > 0 \}$, be the set of indices for all image pixels in the anchor image with overlapping pixels in image V. Finally, we define $\phi_{\mathrm{sem}} \equiv \{ \phi \}^{H_{\mathrm{img}} \times W_{\mathrm{img}} \times K}$ to be the set of image features, $d_{\mathrm{bin}} \equiv \{ 0, ..., C-1 \}^{H_{\mathrm{img}} \times W_{\mathrm{img}} \times 1}$ to be the binned depth label, and $d \equiv \{\mathbb{R}\}^{H_{\mathrm{img}} \times W_{\mathrm{img}} \times 1}$ to be the continuous depth label.

We use superscripts to distinguish which image the variable is associated with and $\hat{}$ symbols to denote variables predicted by our model. 

1. \textbf{Multiview consistency loss} ($\mathcal{L}_{\mathrm{mv}}$): The multiview consistency loss minimizes differences between predicted features for the same BEV patch in the world from different views. We minimize the $l2$ loss between predicted features from opposing views with the anchor view.

\begin{equation}
\mathcal{L}_\mathrm{{mv}} =  \sum_{i \in I^{\mathrm{anchor}}} \sum_{v \in V} \sum_{b \in B} ||\hat{\phi}^{\mathrm{anchor}}_{\mathrm{sem}}(i) - \hat{\phi}^{v}_{\mathrm{sem}}(b)||^2_2
\end{equation}

2. \textbf{Foundation model loss} ($\mathcal{L}_{\mathrm{fdn}}$): The foundation model loss minimizes differences between predicted features and the image feature extracted by a foundation model. In our case, we minimize the $l2$ loss with respect to Dinov2~\cite{oquab2023dinov2} features. We use PCA to reduce each Dinov2 feature to a 64-dimensional vector and normalize to a unit hypersphere before computing the loss.

\begin{equation}
\mathcal{L}_{\mathrm{fdn}} = \frac{1}{|I^{\mathrm{anchor}}|} \sum_{i \in I^{\mathrm{anchor}}} ||\hat{\phi}_{\mathrm{sem}}(i)- \phi_{\mathrm{sem}}(i)|| 
\end{equation}

3. \textbf{Depth classification and regression loss} ($\mathcal{L}_{\mathrm{depth}}$): The depth loss minimizes differences between the predicted depth and the ground truth depth label. We sum cross entropy $\mathcal{L_{CE}}$ and $l1$ losses to form our full depth loss. In practice, we use 128 bins for C, setting the 0 label to invalid or unknown. We find empirically that this offers better performance than using either loss term alone.

\begin{equation}
\mathcal{L}_{\mathrm{depth}} = \frac{1}{|I^{\mathrm{anchor}}|} \sum_{i \in I^{\mathrm{anchor}}} |\hat{d}(i) - d(i)|^1_1 + \mathcal{L_{CE}}( \hat{d}_{\mathrm{bin}}(i), d_{\mathrm{bin}}(i) )
\end{equation}